\documentclass{article}

\usepackage{arxiv}

\usepackage[utf8]{inputenc} 
\usepackage[T1]{fontenc}    
\usepackage{hyperref}       
\usepackage{url}            
\usepackage{booktabs}       
\usepackage{amsfonts}       
\usepackage{nicefrac}       
\usepackage{microtype}      
\usepackage{lipsum}		
\usepackage{graphicx}
\usepackage{natbib}
\usepackage{doi}

\usepackage{amsmath}
\usepackage{booktabs}
\usepackage{floatrow}
\newfloatcommand{capbtabbox}{table}[][\FBwidth]
\DeclareMathOperator*{\argmax}{argmax} 
\usepackage{enumitem}
\usepackage[font=small,labelfont=bf]{caption}
\setlist{nosep, leftmargin=14pt}
\usepackage{subcaption}
\usepackage{mwe} 
\usepackage{bbding}
\usepackage{pifont}
\usepackage{wasysym}
\usepackage{amssymb}
\usepackage{pgfplots} 
\usepackage{array}
\usepackage{multirow}
\usepackage[bottom]{footmisc}
\makeatletter
\renewcommand{\fnum@figure}{Figure. \thefigure}
\makeatother
\pgfplotsset{width=6.5cm, compat=1.6} 
\newsavebox\mybox
\newlength\mylength
\newcommand\boxup[2]{%
  \savebox\mybox{#1}%
  \setlength\mylength{\wd\mybox}%
  \parbox{\mylength}{#1 \\ #2}%
}

\title{Augmenting Knowledge Distillation with Peer-to-Peer Mutual Learning for Model Compression}

\author{ \href{https://orcid.org/0000-0002-5478-3149}{\includegraphics[scale=0.06]{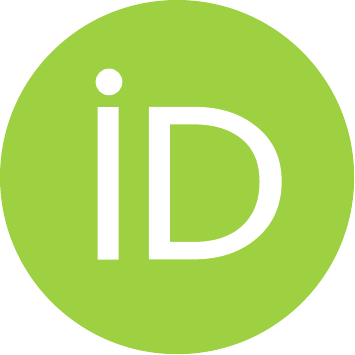}\hspace{1mm}Usma Niyaz} \\
	Department of Computer Science and Engineering\\
	Indian Institute of Technology Ropar\\
	\And \href{https://orcid.org/0000-0002-1383-3744}{\includegraphics[scale=0.06]{orcid.pdf}\hspace{1mm}Deepti R. Bathula} \\
	Department of Computer Science and Engineering\\
	Indian Institute of Technology Ropar\\
}

\date{}

\renewcommand{\headeright}{}
\renewcommand{\shorttitle}{Augmenting Knowledge Distillation with Peer-to-Peer Mutual Learning for Model Compression}

\hypersetup{
pdftitle={A template for the arxiv style},
pdfsubject={q-bio.NC, q-bio.QM},
pdfauthor={David S.~Hippocampus, Elias D.~Striatum},
pdfkeywords={First keyword, Second keyword, More},
}

\begin{document}
\maketitle

\begin{abstract}
Knowledge distillation (KD) is an effective model compression technique where a compact student network is taught to mimic the behavior of a complex and highly trained teacher network. In contrast, Mutual Learning (ML) provides an alternative strategy where multiple simple student networks benefit from sharing knowledge, even in the absence of a powerful but static teacher network. Motivated by these findings, we propose a single-teacher, multi-student framework that leverages both KD and ML to achieve better performance. Furthermore, an online distillation strategy is utilized
to train the teacher and students simultaneously. To evaluate the performance of the proposed approach, extensive experiments were conducted using three different versions of teacher-student networks on benchmark biomedical classification (MSI vs. MSS) and object detection (Polyp Detection) tasks. Ensemble of student networks trained in the proposed manner achieved better results than the ensemble of students
trained using KD or ML individually, establishing the benefit of augmenting knowledge transfer from teacher to students with peer-to-peer learning between students.
\end{abstract}


\section{Introduction}
Deep learning has attracted significant interest from the health care sector over the last decade. With assistance from medical experts and researchers, it has achieved promising results in several areas such as drug discovery, medical image analysis, robotic surgeries, etc.  While deep learning techniques perform exceptionally well, they require training huge models on large datasets to achieve this feat. However, huge models are not always practically feasible due to computational costs associated with training them and their inability to scale to smaller devices. Consequently, there is a shift towards developing deep learning models that are smaller, faster and more efficient that do not compromise performance significantly. Recently, knowledge distillation (KD) has emerged a potential candidate for creating such smaller and efficient architectures. It involves transfer of knowledge gained by a huge pre-trained teacher network to a compact student model. The teacher-student interaction mechanism is designed to gradually enable the smaller student to replicate the behavior of highly trained teacher model.
Knowledge distillation is now considered an established and effective model compression technique (\cite{C1}; \cite{C2}; \cite{C3}). Its applications include a wide variety of computer vision tasks including segmentation (\cite{C4}), object detection and recognition (\cite{C5}). Since its inception, several variants of KD have been proposed to enhance knowledge transfer. Traditionally, the teacher-student networks are trained in a two-stage process called offline KD. The pre-trained teacher network remains fixed while providing structured knowledge to guide the learning process of the student. In contrast, online distillation (\cite{C8}) considers all networks as peers and collaboratively trains them in a single step process. Deep Mutual Learning (DML) (\cite{C7}) achieves promising results by distilling information of logits between the two student networks. Generally, transfer of knowledge is achieved using logits but transfer of intermediate level representations has been explored as well. FitNet (\cite{C6}) transfers feature maps from a pre-trained teacher network to improve the supervised learning of a student network. Furthermore, ensembling of logits produced by all the students has shown to outperform methods that use the logits information directly (\cite{C8}). Evolutionary distillation is proposed in \cite{C9} and \cite{C10} that transfers the intermediate level representation by introducing guided modules in between the teacher and the student network. In addition to computer vision, benefits of KD have also been leveraged in medical image analysis. Mutual Knowledge Distillation (MKD) was proposed to transfer the knowledge from one modality (MR images) to another (CT images) for segmentation task (\cite{C4}). Further in \cite{C11}, knowledge is distilled from the teacher network trained on multimodal data to the single modalilty student network for the Alzheimer's disease prediction.\\
\indent Taking inspiration from classroom learning dynamics, we explore the idea of augmenting the knowledge distillation from teacher to student with mutual learning between multiple students. Our main contributions are:
\begin{itemize}
	\item We propose to combine the benefits of knowledge distillation with mutual learning using a single-teacher, multi-student framework.
	\item Our online training framework involves passing the predictions from the teacher to each of the students and simultaneously sharing of logits information between the students at each iteration.
	\item We demonstrate the efficacy of our proposed approach on benchmark biomedical classification and detection tasks using three different network configurations.
\end{itemize}

\begin{figure*}[ht]
	\centering
	\includegraphics[width=0.90\textwidth,height=8cm]{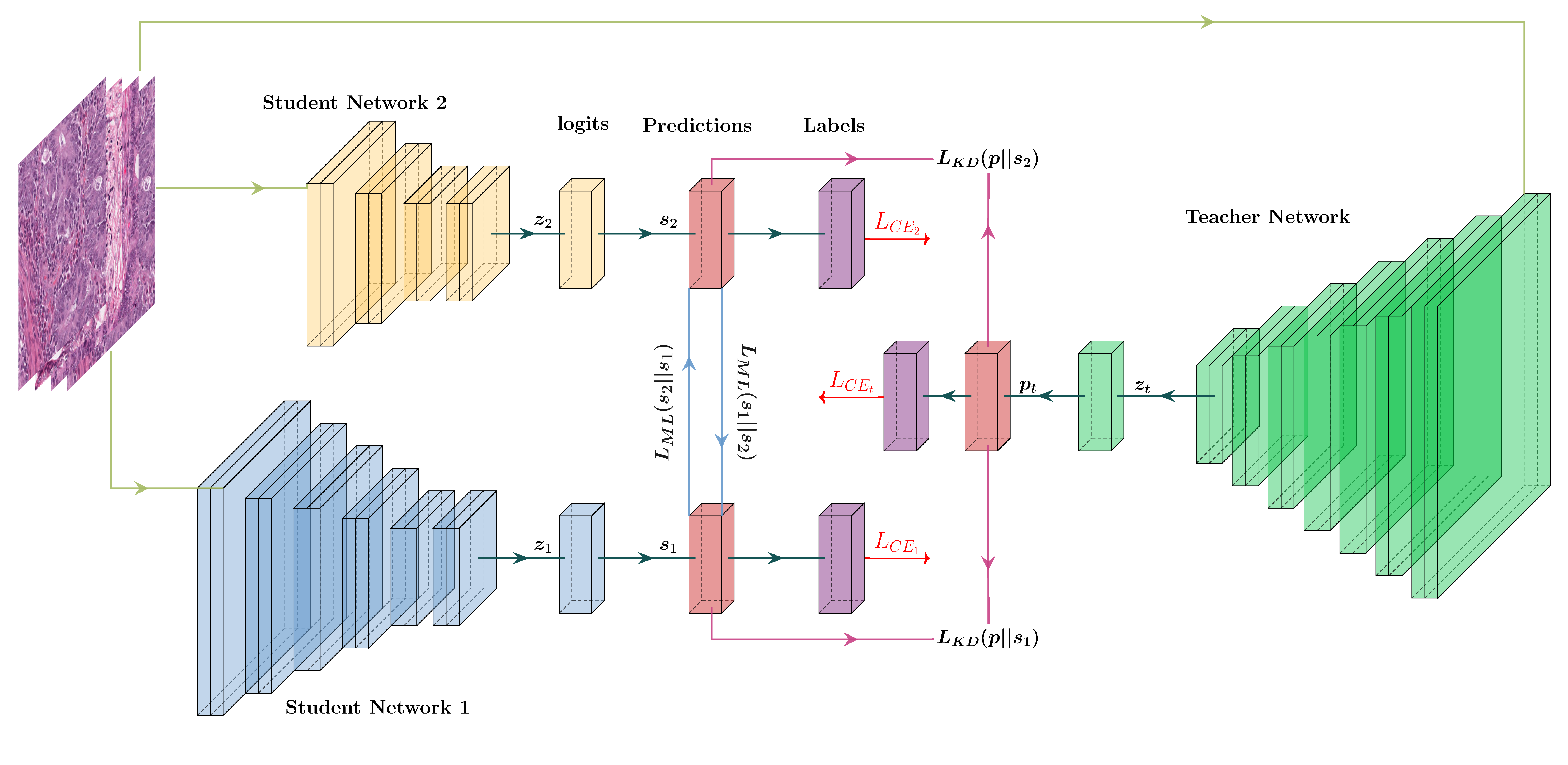}
	\caption{Overview of the combined knowledge distillation and mutual learning technique with one teacher and two student networks. In addition to the supervised learning loss ($L_{CE}$), each student is trained to mimic the teacher ($L_{KD}$) and learn from its peers ($L_{ML}$).}
	\label{fig:fig1}
\end{figure*}

\section{Methodology}
\label{sec:headings}

We formulate the proposed approach using one-teacher and two-student networks as depicted in Figure \ref{fig:fig1}. Extension to more student networks is straightforward as described below. Given a training set with $N$ samples 
${X}=\left\{\boldsymbol{x}_{i}\right\}_{i=1}^{N}$ from $C$ classes with corresponding labels ${Y}=\left\{y_{i} \right\}_{i=1}^{N}$
with $y_{i} \in \{1,2,...,C\}$, we define our teacher and student models as $f(X,\theta)$ and $f(X,\phi_{k})$ respectively and $k \in \{1,2,...,K\}$ is the number of students. The probability of a sample $\boldsymbol{x}_{i}$ belonging to class $c$ according to a network is given by the extended softmax as:

\begin{equation}
p^{c}\left(\boldsymbol{x}_{i}\right)=\frac{\exp \left(\mathbf{z^{c}}/T\right)}{\sum_{c=1}^{C} \exp \left(\mathbf{z^{c}}/T\right)}
\end{equation}

\noindent where $T$ is the temperature parameter and the logit $\mathbf{z^{c}}$ is the output from the network.

Generally, for a  classification task, \textit{Cross-Entropy (CE)} error between the predicted and true labels is used as the objective function to a train network. Consequently, the teacher network is trained using loss function -

\begin{equation}
L_{CE(t)}=-\sum_{i=1}^{N} \sum_{c=1}^{C} I\left(y_{i}, c\right) \log \left(\hat{y_{i}}^{c}\left(\boldsymbol{x}_{i}\right)\right)
\end{equation}

\noindent where $I$ is an indicator function 
with $I\left(y_{i}, c\right) = 1$ if $y_{i} = c$, otherwise $I\left(y_{i}, c\right) = 0$.

As the goal of our approach is to train the student networks to mimic the performance of the teacher network, a \textit{Knowledge Distillation (KD)} term is added to loss function. Furthermore, students are encouraged to share useful information by introducing a \textit{Mutual Learning (ML)} term to the loss function. Here, Kullback Leibler (KL) divergence is used to measure the correspondence between two network's predictions, both teacher-student and student-student. 

Assuming the teacher and student network's predictions are represented by $\mathbf{p}$ and $\mathbf{s}_{k}$ for $k^{th}$ student respectively, the KD and ML losses are given as $L_{KD} = D_{KL}(\mathbf{p} || \mathbf{s}_{k})$ and $L_{ML} = D_{KL}(\mathbf{s}_{k} || \mathbf{s}_{k'})$, where the KL distance from $\mathbf{p}$ to $\mathbf{q}$ is computed as:
\begin{equation}
D_{KL}\left(\boldsymbol{p} \| \boldsymbol{q}\right)=\sum_{i=1}^{N} \sum_{c=1}^{C} p^{c}\left(\boldsymbol{x}_{i}\right) \log \frac{p^{c}\left(\boldsymbol{x}_{i}\right)}{q^{c}\left(\boldsymbol{x}_{i}\right)}
\end{equation}

Consequently, the combined loss function for training the $k^{th}$ student network is:
\begin{equation}
L_{k}= \alpha \ L_{CE(k)} + \beta \ L_{KD}(\mathbf{p}, \mathbf{s}_{k}) + \gamma \  \sum_{\substack{k^{'}=1 \\ k^{'}\neq k}}^K  L_{ML}(\mathbf{s}_{k}, \mathbf{s}_{k^{'}})
\end{equation}

 The ensemble class probabilities $\mathbf{\hat{s}}$ of the multi-student network for the classification task is defined as $\mathbf{\hat{s}}=max\{\mathbf{{s}_k}\}_{k=1}^{K}$  and final label prediction $\mathbf{\hat{y}} = \argmax_{c \in \mathcal{C}} \mathbf{\hat{s}}$.

 
\section{Experimental Results}
\subsection{Datasets}
We evaluated the performance of the proposed approach on both biomedical classification and object detection tasks. For classification, we chose TCGA-COAD MSI vs MSS Dataset\footnote{https://www.kaggle.com/joangibert/tcga\_coad\_msi\_mss\_jpg} for identifying the type of gastrointestinal cancer. Due to limited computational resources, we randomly chose a subset of 5000 histological images from the 192K image repository, ensuring that sample has the same distribution of the population. For or object detection, we chose  Hyper Kvasir Dataset\footnote{https://www.kaggle.com/kelkalot/the-hyper-kvasir-dataset} consisting of 1000 segmented images for polyps detection.
\begin{table}[ht]
	\caption{Different versions of teacher-student networks with compact models for students and more complex networks for teacher.} 
	\centering
	\resizebox{8cm}{!}{
		\begin{tabular}{|l|c|c|c|}
			\hline Versions & V1 &V2 &V3\\
			\hline Teacher Network & Resnet152 & Resnet152 & Densenet201  \\
			Student Network 1 & Resnet50 &  VGG16 &  EfficientNetB0  \\
			Student Network 2 & Resnet34 & VGG19 &  EfficientNetB1  \\
			\hline
		\end{tabular}}
		\label{tab:1}
	\end{table}

\subsection{Experiments}
We compared the performance of our combined knowledge distillation and mutual learning approach with baseline KD and ML only models. Additionally, we trained our proposed model using both online and offline distillation strategies. All models used one-teacher and two-student networks (Figure. \ref{fig:fig1}) and were compared across three different configurations as shown in Table \ref{tab:1}. The Datasets were split for training and testing with 80:20 ratio and experiments were repeated with different random seed values to generalize the networks and ensure reproducibility. The SGD optimizer was used for training the models for classification task. We set the learning rate to $0.01$ with momentum of $0.9$, batch size as $16$ and the values of $T$, $\alpha$, $\beta$ and $\gamma$ are set to $2$, $0.1$, $0.45$ and $0.45$ respectively. During the training process, $L2$ regularization and simple data augmentation techniques are used to reduce the effects of overfitting.

\begin{figure}[ht]
\resizebox{7cm}{!}{
	\begin{tikzpicture} 
	\begin{axis}
	[  
	ybar, 
	enlargelimits=0.15,
	legend cell align=left,
	legend style={at={(0.75,0.95)}, 
		anchor=north,legend columns=1},
	width=10cm, height=5.5cm,     
	ylabel={\#Parameters(millions)}, 
	symbolic x coords={V1, V2, V3},  
	xtick=data,  
	nodes near coords,
	nodes near coords align={vertical},  
	legend image code/.code={
        \draw [#1] (0cm,-0.1cm) rectangle (0.2cm,0.25cm); },
	]  
	\addplot coordinates {(V1, 70) (V2, 70) (V3, 21)}; 
	\addplot coordinates {(V1, 49) (V2, 37) (V3, 14)};  
	\legend{Teacher, Students (combined)}  
	
	\end{axis}  
	
	\end{tikzpicture}}
	\caption{Model Compression: Reduction in parameters between teacher and combined student networks of approximately 30\%  in V1,  47\% in V2 and 33\% in V3 configurations. }
	\label{fig:M1}
\end{figure}
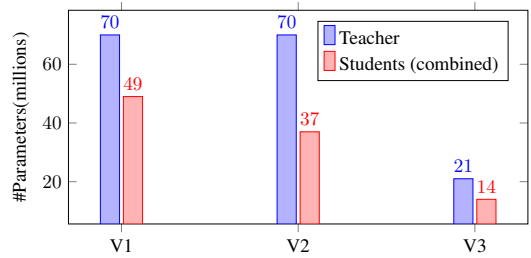

\subsection{Results}
\begin{figure*}[ht]
	\centering
	\includegraphics[width=0.90\textwidth,height=3.5cm]{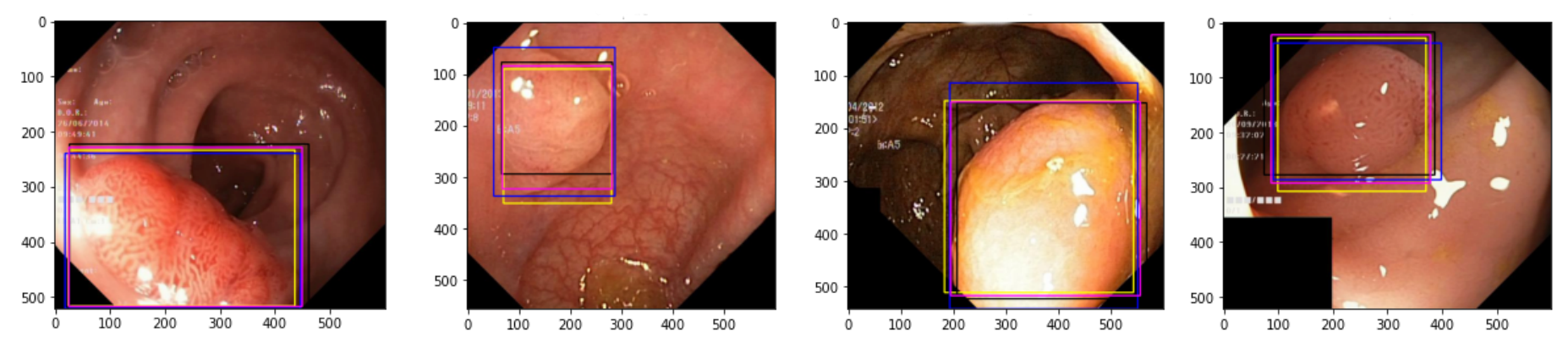}
	\caption{Qualitative comparison: Detection of abnormal polyp tissue with version V2. Bonding boxes: \textit{Blue} – ground truth, \textit{Yellow} – student network VGG16, \textit{Black} - student network VGG19 and \textit{Magenta} – student ensemble.}
	\label{fig:M3}
\end{figure*}

Quantitative results are shown in Table \ref {tab:2}. It can be observed that our proposed model achieves statistically significant improvement in performance than baseline models for both MSI vs. MSS classification and Polyp Detection tasks. This is also true for all three versions of teacher-student network configurations we experimented with. In general, it was also noticed that online distillation that involved joint training of teacher and student networks performed better than the two-stage offline distillation method. The only exception being polyp detection with version V3 where an individual student’s performance in offline and combined model performed better. However, the ensemble IoU value was still better than the baseline KD performance. Qualitative comparison of the models for polyp detection is depicted in Figure \ref{fig:M3}. Bounding box predicted by the ensemble of students from the proposed approach is closer to the ground truth than the individual student networks.

\begin{table*}[ht]
		\caption{Performance comparison: Classification accuracy for MSI vs. MSS and IoU for Polyp detection datasets across methods: KD – Knowledge Distillation, ML – Mutual Learning, KD + ML – Combined, On – online training, Off – Offline training, using different versions of teacher-student network configurations as shown in Table \ref{tab:1}. (\textit{* indicates statistically significant improvement over baseline KD ensemble)}}
	\begin{subtable}{\textwidth}
		\centering
		\resizebox{\textwidth}{!}{
		\begin{tabular}{|l|cccc||cccc|}
			\hline
			\multirow{2}{*}{\textbf{\boxup{Version} { V1}} } &  \multicolumn{4}{c||}{MSI vs MSS Classification} & \multicolumn{4}{c|}{Polyp Detection}\\
			\cmidrule{2-9} & KD  & ML  & KD + ML (On) & KD + ML (Off) & KD & ML & KD + ML (On) & KD + ML (Off) \\
			\hline
			\hline
			Teacher & $76.12\pm1.47$ & $-$ & $75.12\pm1.13$   & $76.12\pm1.47$ & $84.45 \pm 4.09$  & $-$ &  $86.63\pm0.67$ & $84.45 \pm 4.09$\\
			\hline
			\hline
			Student 1 & $74.48\pm1.19$ & $73.68\pm1.74$ & $74.52\pm1.64$ & $74.65\pm2.55$ & $80.68\pm0.87$ & $81.91\pm0.66$  & $83.23\pm0.41$ & $81.05\pm0.66$ \\
			
			Student 2 & $71.23\pm1.68$ & $72.99\pm0.41$ & $72.98\pm2.34$ & $71.49\pm0.65$&  $81.38\pm 0.75$ & $83.71 \pm0.87 $  & $83.51\pm0.65$ &  $81.80\pm0.98$\\
		Ensemble & $75.11\pm1.48$\ & $74.33\pm0.99$ & {\boldmath$75.67^{*}\pm0.51$} & $75.30^{*}\pm0.51$ &$82.01\pm1.01$ &$83.75\pm0.80 $ & {\boldmath$84.20^{*}\pm0.89$} &$82.64^{*}\pm0.83$\\
			
		
			\hline
			
		\end{tabular}}
	\end{subtable}
    \vspace{2.5mm}

	\begin{subtable}{\textwidth}
	    
		\centering
		\resizebox{\textwidth}{!}{
		\begin{tabular}{|l|cccc||cccc|}
			\hline
			\multirow{2}{*}{\textbf{\boxup{Version} { V2}} } &  \multicolumn{4}{c||}{MSI vs MSS Classification} & \multicolumn{4}{c|}{Polyp Detection}\\
			\cmidrule{2-9} & KD  & ML  & KD + ML (On) & KD + ML (Off) & KD & ML & KD + ML (On) & KD + ML (Off) \\
			\hline
			\hline
			Teacher & $76.12\pm1.47$ & $-$ & $75.73\pm0.73$   & $76.12\pm1.47$ & $84.45\pm4.09$  & $-$ &   $86.25\pm0.76$ & $84.45\pm4.09$\\
			\hline
			\hline
			Student 1 & $73.11\pm1.49$ & $73.15\pm2.54$  & $74.83\pm1.50$ & $73.71\pm1.92$ & $83.01 \pm 0.63$ & $ 84.42\pm0.66$ & $84.50\pm0.43$  & $84.16\pm0.67$\\
			
			Student 2  & $71.88\pm1.68$ & $74.25\pm1.58$  & $73.71\pm1.34$  & $72.01\pm2.79$ & $83.59  \pm1.32$ & $ 84.56\pm0.86$  & $84.68\pm0.45$ &  $84.41\pm0.52$\\
		Ensemble & $74.26\pm0.45$ &$74.87\pm2.16$  & {\boldmath$75.48^{*}\pm1.04$} 	& $74.92^{*}\pm1.74$ & $84.36\pm0.57$ & $ 84.83\pm0.65$ & {\boldmath$85.43^{*}\pm0.36$} & $85.05^{*}\pm0.85$\\
			
		
			\hline
			
		\end{tabular}}

	\end{subtable}
    \vspace{2.5mm}

    \begin{subtable}{\textwidth}
		\centering
		\resizebox{\textwidth}{!}{
		\begin{tabular}{|l|cccc||cccc|}
			\hline
			\multirow{2}{*}{\textbf{\boxup{Version} { V3}} }&  \multicolumn{4}{c||}{MSI vs MSS Classification} & \multicolumn{4}{c|}{Polyp Detection}\\
			\cmidrule{2-9} & KD  & ML  & KD + ML (On) & KD + ML (Off) & KD & ML & KD + ML (On) & KD + ML (Off) \\
			\hline
			\hline
			Teacher  & $77.67\pm1.05$  & $-$ &  $77.61\pm0.94$   & $77.67\pm1.05$ & $77.61 \pm0.96$ & $-$ &   $81.31\pm0.81$ & $77.61 \pm0.96$\\
			\hline
			\hline
			Student 1 & $72.20\pm1.04$ &$71.26\pm1.50$ & $72.28\pm1.10$ &  $70.60\pm1.61$ & $70.84 \pm4.27$ & $ 71.08\pm4.14$ & $71.88\pm3.92$   &  $72.52\pm 3.15$ \\
			
			Student 2   & $73.12\pm1.07$ & $70.83\pm1.92$  & $72.00\pm1.38$  & $72.50\pm1.05$ & $71.50\pm5.20$ & $72.03\pm6.97$  & $72.87\pm4.07$ &  {\boldmath$75.66\pm4.75$}\\
		Ensemble & $72.78\pm0.69$ & $71.59\pm1.94$  & {\boldmath$73.17^{*}\pm0.66$} 	& $72.58\pm1.23$ & $72.28\pm3.62$ & $ 71.85\pm4.84$ &  $73.47^{*}\pm2.93$ &  $75.07^{*}\pm1.99$\\
			
		
			\hline
			
		\end{tabular}}
	\end{subtable}
	\label{tab:2}
\end{table*}

\subsubsection{Conclusion}
In this work, we proposed to augment the knowledge distillation from a complex teacher to a compact student with mutual learning between multiple simple students. We conducted extensive experiments with three different network configurations of various capacity on benchmark biomedical classification and object detection tasks. Our results demonstrate the effectiveness of the proposed approach with statistically significant improvement, in both classification accuracy and IoU for object detection, as compared to baseline knowledge distillation and mutual learning approaches. Our current efforts were limited to one teacher and two student networks. However, extension to more student networks is straightforward and will be explored to further improve performance.

\bibliographystyle{unsrtnat}
\bibliography{AKDML}  






\end{document}